\documentclass[runningheads]{llncs}
\usepackage[T1]{fontenc}
\usepackage{graphicx}
\usepackage{booktabs}
\usepackage{multirow}
\usepackage{makecell}
\usepackage{amsmath}
\usepackage{amssymb}
\usepackage{subcaption}
\usepackage[numbers]{natbib}
\usepackage{hyperref}
\usepackage{url}
\usepackage{xcolor}
\usepackage{lmodern}
\usepackage{microtype}
\usepackage{placeins}

\graphicspath{{mathvision-benchmark/paper/figures/}{figures/}{./}}

\begin{document}

\title{Math-Vision Diagrams: A Benchmark for Evaluating LLM Mathematical Diagram Generation Capabilities}

\titlerunning{Math-Vision Diagrams: LLM Math Diagram Generation Benchmark}

\author{Harish Kashyap \and
Kiran Byadarhaly \and
Sriram Chakaravarthy \and
Sanyukta Tuti \and
Aryan Mistry}

\authorrunning{Kashyap et al.}

\institute{Pandita AI Inc., 2540 N First St, San Jose CA 95131, USA \\
\email{\{harish,kiran,sriram,sanyukta,aryan\}@pandita.ai}}

\maketitle

\begin{abstract}
The generation of mathematically precise diagrams from textual prompts has emerged as a critical yet underexplored capability of Large Language Models (LLMs). This has been of interest to researchers in the areas of curriculum preparation, automated ranking of problem sets, and scientific publishing. For LLMs to achieve this, it requires perfect coordination between Spatial Reasoning, Mathematical Reasoning, and Rendering systems.
While existing benchmarks such as MathVision, MathVista are built for Math Reasoning or DiagramGenBenchmark, MermaidSeqBench on general purpose diagram generation, no prior work provides a standardized set of prompt, image pairs that can be used to evaluate the LLMs specifically on math diagram generation. This includes fields that span both both text-to-code and text-to-image paradigms. We introduce \textbf{Math-Vision Diagrams}, the first benchmark specifically designed to evaluate LLMs on mathematical diagram generation, and the first to assess text-to-code and text-to-image generation paradigms together in a single unified setting, agnostic of the underlying coding language or model type. Building on the Math-Vision benchmark, we select a subset of 2920 images out of 3040 from high-quality competition problems with essential visual context. A novel pipeline combining an ensemble of LLMs with Subject Matter Expert (SME) curation is presented, together with a suite of evaluation metrics. Testing several leading models against this benchmark, we demonstrate that LLMs struggle with math diagram generation. All code, data, curation pipeline, and evaluation scripts will be fully open-sourced.

\end{abstract}

\section{Introduction}

The generation of mathematical diagrams from textual descriptions has emerged as an increasingly important research problem at the intersection of multimodal capabilities, structured visual representation, and automated ranking of problem sets. Unlike general text-to-image synthesis, this task requires not only visual plausibility but also mathematical precision. Mathematical diagrams often encode exact spatial relations, labels, and geometric constraints that must be satisfied for the diagram to be pedagogically and mathematically valid. A single misplaced label, incorrect angle, or inconsistent placement of geometric figures can render an entire diagram useless or misleading. Recent advances in Large Language Models (LLMs) have substantially improved performance on tasks that require joint reasoning over text, symbols, and images \cite{lu2023mathvista,chen2021geoqa}.

LLMs such as GPT-4o and Opus now demonstrate impressive capabilities in visual question answering, chart understanding, and basic multimodal mathematical reasoning. Diffusion models have traditionally performed poorly on such visualizations, as they focus on photorealism rather than structured rendering. Prior work has largely focused on \emph{understanding} rather than proving generation ability on a comprehensive image set. For instance, LLMs reasonably interpret diagrams on benchmarks such as MathVista and MathVision, but perform far worse on the converse: generating the same diagram from a pure textual problem statement. This distinction is consequential.

A close examination of the recent literature reveals three key limitations in the current landscape. First, the majority existing benchmarks focus on evaluating \emph{multimodal mathematical reasoning} involving textual descriptions paired with visual input, rather than addressing the generation of mathematical diagrams. Benchmarks such as MathVista, MATH-Vision, MathVerse, and MV-MATH primarily assess reasoning performance over text-image pairs and visual math problems, highlighting a strong emphasis on reasoning rather than structured visual generation~\cite{mathvista2023, mathvision2024, mathverse2024, mvmath2025}. Second, existing datasets and solvers for geometric reasoning are largely concentrated on \emph{plane geometry and question answering tasks}. For instance, GeoQA formalizes geometric reasoning as a question answering problem, while MathVerse categorizes tasks primarily into plane geometry, solid geometry, and functions~\cite{geoqa2021, mathverse2024}. This focus has enabled progress in classical geometry reasoning but leaves broader diagrammatic categories such as transformation diagrams, graph-theoretic structures, and theoretical math schematics relatively underexplored. Third, standard text-to-image generation pipelines are not well suited for mathematical diagrams. Unlike natural images, mathematical diagrams require \emph{discrete structure, symbolic fidelity, precise spatial constraints, and mathematical reasoning}.

Recent work on text-to-diagram generation highlights that outputs from conventional text-to-image systems are often unstructured and difficult to modify~\cite{text2diagram2025}. Similarly, R2I-Bench demonstrates that even advanced text-to-image models exhibit limited reasoning capabilities when faced with structured visual tasks~\cite{r2ibench2025}. These findings suggest that mathematical diagram generation is fundamentally a problem of \emph{structured visual synthesis}, requiring a more complex workflow than a simple extension of photorealistic image generation.

In this paper, we study \emph{math diagram generation} as a first-class problem in multimodal AI. We define the task as mapping a textual or formal mathematical specification to a diagram that are simultaneously :
\begin{enumerate}
    \item Visually interpretable,
    \item Semantically faithful to the source specification,
    \item Mathematically valid under explicit constraints.
\end{enumerate}

These challenges make mathematical diagram generation a particularly revealing testbed for structured multimodal intelligence. Advancing this area may benefit not only mathematics education, but also diagram-centric reasoning in engineering, architecture, data communication, and scientific visualization.

This benchmark is the first of its kind in the field of math diagram generation. By evaluating both text-to-code LLMs (e.g., generating executable TikZ/SVG code) and text-to-image models (including diffusion models and closed-source systems such as Nano Banana Pro), we expose fundamental trade-offs: symbolic precision versus perceptual realism.

The importance of such a benchmark cannot be overstated. Flawed diagrams can mislead learners in educational settings or undermine credibility in research works. Math-Vision Diagrams provides a standardized, reproducible testbed that drives innovation toward human-level mathematical visual intelligence.

\section{Related Work}

\subsection{Previous Work on Diagram Generation}

There has been increasing interest in diagram generation across disciplines such as mathematics, computer science, and physics. General text-to-diagram tasks are formalized in DiagramGenBenchmark \cite{wei2025diagramgen} (CVPR 2025), which introduces eight categories of diagrams (flowcharts, mind maps, architecture diagrams) and evaluates code-based outputs via ROUGE-L, codeBLEU, and compilation rates. The benchmark highlights the limitations of prior text-to-image approaches when applied to structured visuals, emphasizing the need for hybrid evaluation protocols that capture both syntactic validity and semantic fidelity.

SVGEditBench \cite{nishina2024svgeditbench} and MathemaTikZ \cite{malik2025mathematikz} (L@S 2025) target SVG/TikZ generation for symbolic graphics and K-12 mathematical diagrams. These works demonstrate that vector-based outputs offer superior editability and precision compared to raster images, but they remain limited in scale and mathematical depth. Our benchmark differs from MathemaTikZ in three ways: it draws on competition-grade MathVision sources rather than K-12 material, spans 16 mathematical disciplines rather than a narrower curricular slice, and evaluates text-to-code and text-to-image paradigms jointly under a shared metric suite. MermaidSeqBench \cite{shbita2025mermaidseqbench} (NeurIPS 2025 workshop) benchmarks LLM-to-Mermaid sequence diagrams, further illustrating the growing interest in domain-specific diagram languages.

Math-specific and scientific diagram generation has seen significant progress. \textit{DeTikZify: Synthesizing Graphics Programs for Scientific Figures and Sketches with TikZ} \cite{belouadi2024detikzify} (NeurIPS 2024 spotlight) trains multimodal models on over 360K TikZ examples for semantics-preserving scientific figure synthesis. The authors show that large-scale pre-training on vector graphics significantly improves structural accuracy, but note persistent challenges with complex geometric constraints. \textit{From Text to Visuals: Using LLMs to Generate Math Diagrams with Vector Graphics} \cite{lee2025svgmath} (AIED 2025) explores SVG intermediates for educational math hints, reporting improvements in student comprehension when diagrams are generated on-the-fly.

\textit{Draw with Thought: Unleashing Multimodal Reasoning for Scientific Diagram Generation} \cite{cui2025drawwiththought} (ACM MM 2025) proposes a training free Chain-of-Thought framework using mxGraph XML and releases Plot2XML, a benchmark of 247 real-world scientific diagrams. This work underscores the value of structured intermediate representations for maintaining mathematical integrity. \textit{MagicGeo: Training-Free Text-Guided Geometric Diagram Generation} \cite{wang2025magicgeo} (arXiv 2025) introduces coordinate optimization with formal solvers and MagicGeoBench (220 geometric descriptions), achieving state-of-the-art results on plane geometry tasks through constraint satisfaction.

Vector-graphics focused works include \textit{Empowering LLMs to Understand and Generate Complex Vector Graphics} \cite{xing2025llm4svg} (CVPR 2025) with LLM4SVG, \textit{Chat2SVG: Vector Graphics Generation with Large Language Models and Image} \cite{wu2025chat2svg} (CVPR 2025), and \textit{OmniSVG: A Unified Scalable Vector Graphics Generation Model} (NeurIPS 2025). Broader multimodal benchmarks such as MMMG \cite{mmmg2025} (NeurIPS 2025) incorporate diagrams, charts, and mind maps across disciplines. Additional efforts appear in IEEE venues, e.g., \textit{A Precise Text-to-Diagram Generation Method for Elementary Geometry} \cite{zhengyu2023precise} (IEEE ICCWAMTIP 2023) and \textit{DiagramIR: An Automatic Pipeline for Educational Math Diagram Evaluation} \cite{kumar2025diagramir} (NeurIPS Math-AI 2025 workshop).

No prior benchmark unifies math diagram generation across code-based, raster-based and image based paradigms while using a competition-grade source like MathVision. Our work fulfills this gap.

\section{Dataset}

\subsection{Dataset Overview, Development \& Curation}
We introduce Math-Vision Diagrams, a benchmark dataset built using a novel pipeline that couples LLM-based processing with Subject Matter Expert (SME) curation, aimed at evaluating the ability of LLMs to generate mathematical diagrams. We base it on the MathVision dataset \cite{mathvision2024}, a comprehensive multimodal math reasoning benchmark of 3,040 authentic competition problems spanning 16 mathematical disciplines including algebra, geometry (metric, analytic, solid, combinatorial, descriptive, transformation), combinatorics, graph theory, statistics, and topology. We remove purely illustrative images (graffiti, cartoons, other non-math visuals) and retain a subset of 2920 problems that are purely vision-based math reasoning exercises. These diagrams have high visual dependence: they are not merely illustrative but encode essential geometric constraints, with authentic competition-grade complexity.

\subsection{Dataset Development}
We developed a novel, fully reproducible curation pipeline that combines automated vision-language processing with rigorous human SME oversight, a hybrid approach for math diagram benchmark construction. Images such as photos, illustrations, or decorative visuals that are not faithfully reproducible as formal mathematical diagrams are eliminated from the list.

The pipeline consists of four main steps:
\begin{itemize}
    \item \textbf{Step 1: Filtering} The original dataset is filtered to contain only vision based math reasoning tasks. They are further filtered to remove non-mathematical diagrams.
    \item \textbf{Step 2: Multi-VLM Descriptions} Multiple VLMs run in parallel to produce rich visual descriptions. Failed calls are automatically retried and patched in-place to ensure completeness.
    \item \textbf{Step 3: Multimodal Grounded Refinement} A separate open-source judge model aggregates these descriptions with direct access to the source image and resolves conflicts by image-grounded verification
    \item \textbf{Step 4: Downstream validation.} Reliability is validated via description-level quality checks by human Subject Matter Experts.
\end{itemize}

For images retained as diagrams, we then collect three independent model descriptions (GPT family, Gemini family, and Claude family) using a common visual checklist. A separate open-source judge model aggregates these descriptions with direct access to the source image and resolves conflicts by image-based verification. This design reduces circularity because the aggregation model family is not reused as a primary describer.

\subsection{Building Input Prompt, Image Ground Truth Pairs}
\label{sec:stage1}

\subsubsection{Concise Prompt Synthesis.}
\label{sec:synthesis}

Four free-form descriptions per image is unwieldy as a model prompt.
We condense them using \textbf{Llama 3.3-70B} (via Groq API) as a
text-only judge, which reads all available descriptions usually four,
occasionally fewer, and produces a single ``Draw\,$\ldots$'' prompt that is capped.

The judge is instructed to prioritize overall structure, exact labels, text, and spatial layout, and to omit stylistic defaults like white
backgrounds or implicit grid lines. 

\subsection{Prompt, Image Pairs}
The resulting output is a {prompt, image} pair as shown in the example below. The image is the ground truth.

\begin{table}
\label{tab:prompt_image_example}
\centering
\small
\begin{tabular}{p{0.56\textwidth}p{0.38\textwidth}}
\toprule
\textbf{Text description} & \textbf{Reference image} \\
\midrule
\parbox[t]{\linewidth}{Draw a circle with a horizontal and vertical diameter intersecting at the center, labeled ``O''. Include a small rectangle in the lower-right quadrant with a diagonal line, labeled ``5'', and the bottom edge labeled ``4''. Add right-angle markers at the rectangle's corners and the center intersection.}
&
\parbox[c]{\linewidth}{\centering\includegraphics[width=0.9\linewidth]{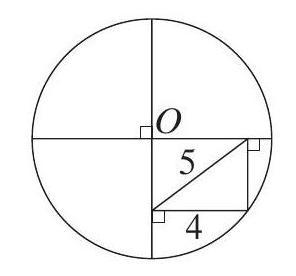}} \\
\bottomrule
\end{tabular}
\end{table}

\subsection{Prompt-Sufficiency Validation}
\label{sec:prompt-sufficiency}
Because each prompt is synthesized from VLM descriptions of the target
image rather than from the original competition problem statement, the
task we measure is closer to \emph{caption-conditioned reconstruction}
than to generation from an authentic mathematical prompt. This is a
deliberate design choice: original problem statements frequently
under-specify the diagram (they assume the figure is already given),
whereas a reconstruction prompt must stand on its own. To verify that
our synthesized prompts are in fact sufficient to recreate the
reference diagram, two Subject Matter Experts (SMEs) with graduate-level
mathematics backgrounds independently scored a random sample of
50 prompt--image pairs on three 1--5 dimensions: \emph{completeness}
(does the prompt mention every element in the image?),
\emph{correctness} (is every statement in the prompt true of the
image?), and \emph{clarity} (could a competent reader redraw the image
from the prompt alone?).

Across both assessors, the mean scores were high on all three axes
(completeness $4.95$, correctness $4.85$, clarity $4.78$; grand mean
$4.86/5$ over $300$ judgements). The exact agreement between the two parties was
$90\%$ for completeness, $90\%$ for correctness, and $80\%$ for
clarity, with agreement within one point on $100\%$ of the items,
indicating that the two SMEs converged closely. Treating a clarity
score of $\geq 4$ as ``sufficient to redraw,'' $47$ of $50$ prompts
($94\%$) met the bar for both raters. The complete scores per-image are 
reported in the Appendix Table~\ref{tab:sme-full}.

The residual ambiguity is concentrated in a few cases and is itself
informative. Table~\ref{tab:sme-examples} contrasts a high-agreement
prompt with a lower-clarity one: the first specifies shape, labels,
shading, and relative sizing unambiguously, whereas the second
describes a tilted arrangement of digit tiles whose exact grouping and
orientation are difficult to pin down in text. We retain such cases
rather than discarding them, but flag prompt clarity as a known source
of variance that downstream users should account for.

\begin{table}[!htb]
\centering
\caption{Representative prompt-sufficiency cases. SME scores are
  (Completeness, Correctness, Clarity) on a 1--5 scale, shown as
  SME~1 / SME~2.}
\label{tab:sme-examples}
\small
\begin{tabular}{p{0.20\textwidth} p{0.46\textwidth} c}
\toprule
\textbf{Reference} & \textbf{Synthesized prompt} & \textbf{SME 1 / 2} \\
\midrule
\raisebox{-\totalheight}{\includegraphics[width=0.18\textwidth]{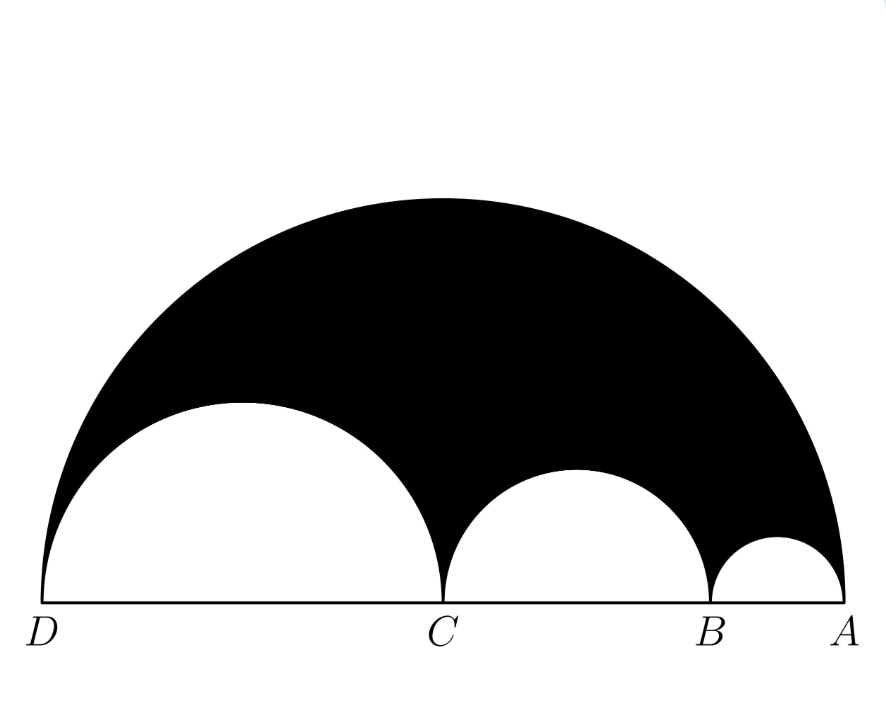}}
&
{\footnotesize Draw a large semicircle with diameter AD, and three
smaller semicircles with diameters DC, CB, and AB, arranged side by
side along the baseline. Label the endpoints D, C, B, and A. Fill the
region inside the large semicircle but outside the three smaller
semicircles with solid black shading, leaving the smaller semicircles
unshaded. DC is the largest of the three, CB medium, AB the smallest.}
&
\makecell{(5,5,5)\\(5,5,5)} \\
\midrule
\raisebox{-\totalheight}{\includegraphics[width=0.16\textwidth]{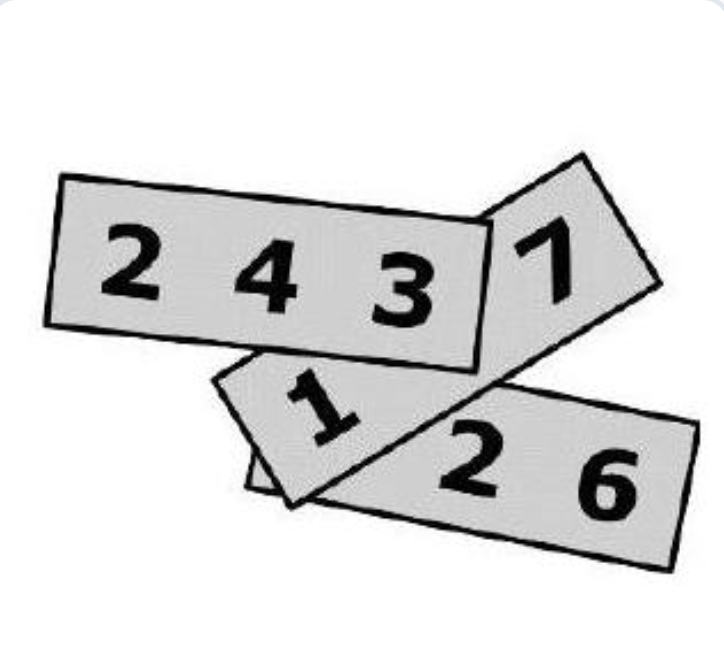}}
&
{\footnotesize Draw three overlapping gray rectangles with black
outlines, each containing digits: top-left ``2 4 3'', top-right ``1'',
and bottom ``2 6''. Position them to suggest a 3D stack, with the top
rectangle overlapping the others.}
&
\makecell{(5,4,3)\\(4,4,4)} \\
\bottomrule
\end{tabular}
\end{table}

\section{Experiments}
\label{sec:experiments}

The prompt--ground-truth-image pairs produced by the curation pipeline described in Section 3, we evaluate eleven models spanning two generation paradigms: text-to-code(compiled to get an image) and text-to-image. This section details the generation and compilation protocol (Section~\ref{sec:exp-setup}), the evaluation metrics and statistical methodology (Section~\ref{sec:exp-metrics}), the quantitative results across all models and diagram categories (Section~\ref{sec:exp-results}), and a qualitative error analysis of representative failure modes (Section~\ref{sec:exp-errors}).

\subsection{Experimental Setup}
\label{sec:exp-setup}

\subsubsection{Models Evaluated.}
\label{sec:exp-models}

We test \textbf{eleven models} across two paradigms (Table~\ref{tab:models}). Nine are code-generating LLMs that produce executable TikZ, SVG, or Python/Matplotlib specifications compiled to raster images. Two are text-to-image generation models that synthesize raster diagrams directly from the textual prompt, bypassing any compilation step. This pairing enables a controlled comparison of symbolic precision versus direct perceptual synthesis on identical inputs and metrics.

On the proprietary side, \textbf{GPT-5.4} (OpenAI) represents frontier chain-of-thought reasoning; \textbf{GPT-OSS-120B} (OpenAI via Groq) is the corresponding open-weight baseline; \textbf{Claude Opus~4.6} (Anthropic) is a strong structured code generator; and \textbf{Gemini~3.1~Pro} (Google) brings native SVG support and a 1M-token context window. Among open-weight models, \textbf{DeepSeek-R1} and \textbf{DeepSeek-V3} (DeepSeek) cover reasoning and efficiency variants; \textbf{Qwen3.5-35B-A3B} (Alibaba) is compact yet competitive; \textbf{Llama~4~Maverick} (Meta) is a multimodal MoE; and \textbf{Kimi~K2.5} (MoonshotAI) broadens organisational coverage. The two image models, \textbf{Nano~Banana~Pro} and \textbf{Nano~Banana~2} (Google), are designed for diagrams and infographics at native 4K resolution.

\begin{table}[t]
  \caption{Models evaluated in the Math-Vision Diagrams benchmark.
    The ``Open'' column lists the applicable licence; ``, '' denotes
    closed-source models.}
  \label{tab:models}
  \centering
  \small
  \begin{tabular}{llllp{4.6cm}}
    \toprule
    \textbf{Model} & \textbf{Developer} & \textbf{Type}
      & \textbf{Open} & \textbf{Key strength} \\
    \midrule
    GPT-5.4           & OpenAI       & Code LLM & ,         & Frontier reasoning \\
    GPT-OSS-120B      & OpenAI/Groq  & Code LLM & Weights    & Open-weight OAI baseline \\
    Claude Opus~4.6   & Anthropic    & Code LLM & ,         & 65.4\% T-Bench; 80.8\% SWE \\
    Gemini~3.1~Pro    & Google       & Code LLM & ,         & 94.3\% GPQA; native SVG; 1M~ctx \\
    DeepSeek-R1       & DeepSeek     & Code LLM & MIT        & 97.3\% MATH-500; CoT traces \\
    DeepSeek-V3       & DeepSeek     & Code LLM & MIT        & 94\% MATH-500; MoE efficiency \\
    Qwen3.5-35B-A3B   & Alibaba      & Code LLM & Apache~2.0 & 3B active; 1M~ctx \\
    Llama~4~Maverick  & Meta         & Code LLM & Llama      & Multimodal MoE; 1M~ctx \\
    Kimi~K2.5         & MoonshotAI   & Code LLM & MIT        & Open-weight frontier reasoning \\
    \midrule
    Nano~Banana~Pro   & Google       & Text-to-Image.\ image & ,  & \#3 Arena; 4K; diagram mode \\
    Nano~Banana~2     & Google       & Text-to-Image.\ image & ,  & \#2 Arena; Flash speed \\
    \bottomrule
  \end{tabular}
\end{table}

\subsection{Evaluation Metrics}
\begin{table}[h]
  \setlength{\abovecaptionskip}{2pt}
  \setlength{\belowcaptionskip}{2pt}
  \vspace{-0.5\baselineskip}
  \caption{Summary of evaluation metrics used in the benchmark.}
  \label{tab:metrics-main}
  \centering
  \small
  \begin{tabular}{llll}
    \toprule
    \textbf{Metric} & \textbf{Level}
      & \textbf{Direction} & \textbf{Implementation} \\
    \midrule
    DISTS                & Per-image    & $\downarrow$ better
      & \texttt{piq.DISTS()} \\
    CLIP cosine sim.     & Per-image    & $\uparrow$ better
      & CLIP ViT-B/32 \\
    Edge IoU             & Per-image    & $\uparrow$ better
      & Canny + binary mask overlap \\
    Edge F1              & Per-image    & $\uparrow$ better
      & Canny + precision/recall \\
    \bottomrule
  \end{tabular}
\end{table}

\subsection{Experimental Results}
\label{sec:exp-results}
We ran all eleven models on the full set of curated prompts. This section reports where models excel, where they struggle, and what patterns emerge across categories, difficulty levels, and generation paradigms.

\subsubsection{Model Profiles.}
\label{sec:exp-radar}

\begin{figure}[!htb]
  \centering
  \includegraphics[width=0.6\textwidth]{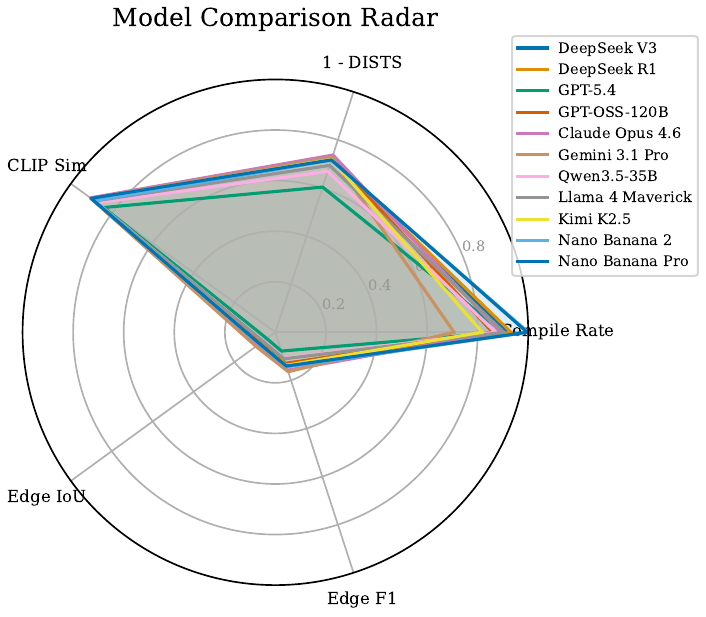}
  \caption{\textbf{Radar chart comparing model profiles across compile rate and four evaluation metrics} on the common subset, with all axes normalized to $[0,1]$. Most models cluster closely on compile rate, CLIP similarity, and $1-\mathrm{DISTS}$, indicating broadly similar perceptual performance, whereas Edge IoU and Edge F1 are substantially lower for all models. This suggests that structural edge fidelity remains the most challenging aspect of the task. The radar plot therefore serves primarily to visualize relative strengths and weaknesses across metrics, rather than to imply a strict overall ranking by polygon area.}
  \label{fig:radar}
\end{figure}

Figure~\ref{fig:radar} summarizes each model's strengths and weaknesses. Claude Opus~4.6 has the most balanced profile, with strong scores on every axis. Gemini~3.1~Pro is stretched toward the structural metrics but pulled inward on compile rate. GPT-5.4's profile is visibly collapsed, confirming that its weakness is systemic rather than confined to a single metric. The Nano~Banana models form a distinctive shape: maximal on reliability, competitive on perceptual quality, but narrower on structural precision.

\subsubsection{Overall Performance.}
\label{sec:exp-overall}
\begin{table}[t]
\centering
\caption{Model performance over all evaluated pairs. Bold = best,
  underline = second best. Values are mean $\pm$ 95\% CI.
  Compile rate is relative to the 2,920 prompts.}
\label{tab:leaderboard_all}
\small
\begin{tabular}{l r r r r r r}
\toprule
Model & $n$ & Compile & DISTS\,$\downarrow$ & CLIP\,$\uparrow$ & Edge IoU\,$\uparrow$ & Edge F1\,$\uparrow$ \\
\midrule
Claude Opus~4.6   & 2690 & 89.1\% & \textbf{0.263\,\scriptsize$\pm$.003} & \textbf{0.905\,\scriptsize$\pm$.002} & \underline{0.089\,\scriptsize$\pm$.003} & \underline{0.158\,\scriptsize$\pm$.004} \\
Gemini~3.1~Pro    & 2083 & 70.7\% & 0.272\,{\scriptsize$\pm$.005} & 0.901\,{\scriptsize$\pm$.004} & \textbf{0.095\,\scriptsize$\pm$.004} & \textbf{0.164\,\scriptsize$\pm$.005} \\
GPT-OSS-120B      & 2582 & 85.9\% & \underline{0.272\,\scriptsize$\pm$.003} & 0.901\,{\scriptsize$\pm$.003} & 0.072\,{\scriptsize$\pm$.002} & 0.129\,{\scriptsize$\pm$.003} \\
Nano Banana Pro   & 2908 & \underline{99.6\%} & 0.283\,{\scriptsize$\pm$.002} & \underline{0.901\,\scriptsize$\pm$.002} & 0.077\,{\scriptsize$\pm$.002} & 0.140\,{\scriptsize$\pm$.003} \\
DeepSeek V3       & 2772 & 91.6\% & 0.273\,{\scriptsize$\pm$.002} & 0.897\,{\scriptsize$\pm$.002} & 0.078\,{\scriptsize$\pm$.002} & 0.140\,{\scriptsize$\pm$.003} \\
DeepSeek R1       & 2808 & 93.2\% & 0.278\,{\scriptsize$\pm$.003} & 0.897\,{\scriptsize$\pm$.002} & 0.073\,{\scriptsize$\pm$.002} & 0.132\,{\scriptsize$\pm$.003} \\
Kimi K2.5         & 2389 & 81.8\% & 0.281\,{\scriptsize$\pm$.004} & 0.898\,{\scriptsize$\pm$.003} & 0.077\,{\scriptsize$\pm$.003} & 0.137\,{\scriptsize$\pm$.004} \\
Nano Banana 2     & 2920 & \textbf{100.0\%} & 0.284\,{\scriptsize$\pm$.002} & 0.886\,{\scriptsize$\pm$.003} & 0.079\,{\scriptsize$\pm$.002} & 0.144\,{\scriptsize$\pm$.003} \\
Llama 4 Maverick  & 2641 & 90.4\% & 0.306\,{\scriptsize$\pm$.004} & 0.884\,{\scriptsize$\pm$.003} & 0.060\,{\scriptsize$\pm$.002} & 0.110\,{\scriptsize$\pm$.003} \\
Qwen3.5-35B       & 2542 & 87.1\% & 0.331\,{\scriptsize$\pm$.005} & 0.872\,{\scriptsize$\pm$.003} & 0.059\,{\scriptsize$\pm$.002} & 0.108\,{\scriptsize$\pm$.003} \\
GPT-5.4           & 2628 & 90.0\% & 0.396\,{\scriptsize$\pm$.005} & 0.839\,{\scriptsize$\pm$.004} & 0.042\,{\scriptsize$\pm$.002} & 0.078\,{\scriptsize$\pm$.003} \\
\bottomrule
\end{tabular}
\end{table}

Table~\ref{tab:leaderboard_all} presents per-model results over every prompt that returned a valid image. \textbf{Claude Opus~4.6 leads on perceptual and semantic metrics}, achieving the lowest DISTS (0.263) and highest CLIP similarity (0.905); its edge metrics are also strong, second only to Gemini~3.1~Pro. \textbf{Gemini~3.1~Pro produces the most structurally precise diagrams when it compiles}, topping Edge~IoU (0.095) and Edge~F1 (0.164), but compiles only 70.7\% of the time, the lowest of any model. \textbf{GPT-5.4 substantially underperforms}, recording the worst DISTS (0.396) and CLIP (0.839) and the lowest edge scores; manual inspection shows it often generates overly complex TikZ with coordinate miscalculations that survive compilation but produce incorrect diagrams. Its open-weight sibling GPT-OSS-120B performs markedly better, suggesting extended chain-of-thought reasoning may hurt code generation. \textbf{DeepSeek~V3 and R1 are strong all-rounders}, with high compile rates ($>$91\%) and top-tier DISTS and CLIP; R1's explicit reasoning traces confer no measurable advantage. The Nano~Banana image models reach near-perfect reliability (100\% and 99.6\%) and handle prompts mixing natural imagery with math diagrams, but their structural precision (Edge~IoU $\approx$~0.077--0.079) still trails the best code generators.

\textbf{Common-subset robustness.} Because compile rates vary widely (70.7\%--100\%), averaging only over successful outputs risks conflating quality with selection effects. We therefore recompute all metrics on the common subset of 1{,}068 prompts that every model rendered successfully (full results in Appendix, Table~\ref{tab:leaderboard_common}). Rankings are stable: Claude Opus~4.6 retains the lead on DISTS (0.245) and CLIP (0.920), Gemini~3.1~Pro still tops both edge metrics, and the gap between these two and the rest \emph{widens} rather than narrows, confirming the advantage is not an artefact of easy prompts.

\section{Conclusion}

We have presented Math-Vision Diagrams, the first dedicated benchmark for evaluating LLM mathematical diagram generation. Our comprehensive evaluation of eleven models reveals three central findings. First, text-to-code LLMs retain a measurable structural advantage over text-to-image models on geometric precision, but this advantage comes at the cost of compilation reliability, the best code models lose 7-29\% of their outputs to compilation failures. Second, no single model dominates all evaluation axes: perceptual quality, structural fidelity, and generation reliability are partially independent properties that must be traded off against each other in application-specific deployment. Third, mathematical subject categories and problem difficulty strongly modulate performance, with standard plane geometry well within reach of current models, while statistical charts, topological figures, and transformation diagrams remain challenging. All code, data, the curation pipeline, evaluation scripts, and leaderboards will be fully open-sourced. 

\begin{credits}
\subsubsection*{Disclosure of Interests} The authors declare no competing interests relevant to this article.
\end{credits}

\bibliographystyle{splncs04}

\newpage

\section*{Appendix}

\section{Diagram Generation by Text-to-Code LLMs}

Text-to-code LLMs generate executable diagram specifications (e.g., TikZ, SVG, or Python/Matplotlib) instead of directly synthesizing pixels. For mathematical diagrams, this paradigm is particularly attractive because it naturally exposes structure: labels, coordinates, constraints, and geometric relations are represented explicitly in code, making outputs easier to verify, edit, and reproduce.

In practice, the pipeline typically follows four steps: (i) prompt a model with a natural-language problem statement, (ii) generate diagram code in a target formalism, (iii) compile/render code to an image, and (iv) evaluate both render quality and structural correctness. This decomposition supports deterministic debugging and allows failures to be localized (e.g., syntax errors, invalid geometry, or semantic mismatch). Recent work on TikZ/SVG-style generation similarly shows that structural representations improve controllability and downstream refinement relative to purely raster generation \cite{belouadi2024detikzify,lee2025svgmath,xing2025llm4svg,wu2025chat2svg}.

A core advantage of text-to-code systems is \emph{verifiability}. Compilability, static checks, and geometry-aware constraints provide concrete pass/fail signals that are difficult to obtain from end-to-end text-to-image models. For example, malformed commands, missing labels, or inconsistent coordinate definitions can be programmatically detected before final rendering. This makes text-to-code approaches well aligned with benchmark settings that prioritize mathematical precision over photorealism.

However, text-to-code generation also introduces distinct challenges. Models may produce syntactically valid code that is semantically incorrect (e.g., wrong angle placement), rely on brittle defaults in plotting libraries, or overfit to common templates that fail on uncommon diagram topologies. Additionally, different target languages (TikZ vs. SVG vs. Matplotlib) expose different expressive biases and failure modes, complicating cross-format comparisons.

In our benchmark design, we therefore treat text-to-code LLMs as a first-class model family and evaluate them jointly with text-to-image systems under shared prompts and metrics. This enables a controlled analysis of the central trade-off: symbolic precision and editability versus direct perceptual realism.

\subsection{Illustrative Prompt-Response Example (Gemini)}
To illustrate text-to-code behavior with the same instruction, consider the prompt: \emph{``draw in latex/xml a circle inscribed in a triangle with all of its sides being a tangent.''} A representative model response may choose Python (Matplotlib + NumPy) to compute and render the incircle.

\paragraph{Prompt}
\texttt{draw in latex/xml a circle inscribed in a triangle with all of its sides being a tangent.}

\paragraph{Representative Python response}
The response below computes side lengths, semi-perimeter, area (Heron's formula), incenter, and inradius; then draws the triangle and its incircle.

\begin{verbatim}
import matplotlib.pyplot as plt
import numpy as np

def draw_incircle():
    # Define triangle vertices
    A = np.array([0, 0])
    B = np.array([5, 0])
    C = np.array([2, 4])

    # Side lengths (a opposite A, b opposite B, c opposite C)
    a = np.linalg.norm(B - C)
    b = np.linalg.norm(A - C)
    c = np.linalg.norm(A - B)

    # Perimeter and semi-perimeter
    p = a + b + c
    s = p / 2

    # Area using Heron's formula
    area = np.sqrt(s * (s - a) * (s - b) * (s - c))

    # Incenter coordinates
    incenter = (a * A + b * B + c * C) / p

    # Inradius
    inradius = area / s

    # Plotting
    fig, ax = plt.subplots()
    triangle = plt.Polygon([A, B, C], fill=None, edgecolor='black', linewidth=2)
    ax.add_patch(triangle)

    circle = plt.Circle(incenter, inradius, fill=None, edgecolor='blue',
                        linewidth=2, label='Incircle')
    ax.add_patch(circle)

    ax.plot(*A, 'ko'); ax.text(A[0], A[1]-0.2, 'A', ha='center')
    ax.plot(*B, 'ko'); ax.text(B[0], B[1]-0.2, 'B', ha='center')
    ax.plot(*C, 'ko'); ax.text(C[0], C[1]+0.1, 'C', ha='center')
    ax.plot(*incenter, 'ro', label='Incenter')

    ax.set_aspect('equal')
    ax.set_xlim(-1, 6)
    ax.set_ylim(-1, 5)
    ax.legend()
    ax.set_title('Circle Inscribed in a Triangle')
    plt.grid(True, linestyle='--', alpha=0.6)
    plt.savefig('inscribed_circle.png')
    plt.close()

draw_incircle()
\end{verbatim}

\paragraph{Geometric formulas used}
Incenter:
\begin{equation*}
I = \frac{aA + bB + cC}{a+b+c}
\end{equation*}
Semi-perimeter:
\begin{equation*}
s = \frac{a+b+c}{2}
\end{equation*}
Inradius:
\begin{equation*}
r = \frac{\text{Area}}{s}
\end{equation*}

This response pattern is useful for benchmarking because it combines symbolic geometry (explicit formulas) with executable rendering logic, enabling both semantic and compilability-oriented evaluation. This example captures the strengths and limits of text-to-code LLMs. The output is a code that is editable, compilable, and structurally interpretable, but correctness still depends on whether geometric constraints are faithfully encoded (e.g., true tangency to all three sides in arbitrary configurations). Our benchmark therefore evaluates not only compilation success, but also semantic geometric validity under multi-metric and SME-grounded assessment.

For reproducibility, Table~\ref{tab:modules} summarizes core modules used in the benchmark pipeline.

\begin{table}
\caption{Compact module reference for the Math-Vision Diagrams pipeline.}\label{tab:modules}
\centering
\begin{tabular}{p{0.23\textwidth}p{0.20\textwidth}p{0.47\textwidth}}
\toprule
\textbf{Module} & \textbf{Layer} & \textbf{Role} \\
\midrule
\texttt{config.py}, \texttt{api\_clients.py} & Infrastructure & Central configuration, provider clients, model routing \\
\texttt{data\_loader.py}, \texttt{utils.py} & Infrastructure & Dataset loading, caching, retries, checkpointing \\
\texttt{taxonomy.py}, \texttt{classification.py} & Classification & Six-category schema and structured boolean classification \\
\texttt{agreement.py}, \texttt{validation.py} & Classification & Reliability scoring, deterministic overrides, downstream checks \\
\texttt{describe.py} & Description & Multi-model description generation with shared checklist \\
\texttt{consensus.py} & Aggregation & Image-grounded conflict resolution and final consolidated description \\
\texttt{report.py} & Reporting & CSV synthesis and interactive qualitative dashboard generation \\
\bottomrule
\end{tabular}
\end{table}

\section{Experimental Setup Details}

\paragraph{Statistical methodology.}
For cross-model comparison, we performed Wilcoxon signed-rank tests on all $\binom{11}{2} = 55$ model pairs for each per-image metric, using the common subset ($n = 1{,}068$) to ensure paired observations. $p$-values are corrected for multiple comparisons using the Holm--Bonferroni step-down procedure at a family-wise error rate of $\alpha = 0.05$. This non-parametric test is appropriate because the per-image score distributions are heavily skewed (particularly Edge~IoU and Edge~F1), violating the normality assumptions required by paired $t$-tests.

\subsubsection{Generation Protocol.}
Each of the eleven models is run independently over all 2,920 prompts. All generation calls use each provider's default temperature (typically $T = 1.0$) with exactly one response per prompt. We do not employ best-of-$k$ sampling or any form of response selection; this single-shot protocol reflects the most common deployment scenario and avoids introducing a selection oracle that would inflate apparent performance.

The nine code-generating LLMs receive a shared system prompt via \texttt{generate.py} that accepts any of TikZ/\LaTeX, SVG, or Python/Matplotlib and requires only a single fenced code block with no surrounding commentary. Leaving the output format unconstrained allows each model to use the formalism it handles best, avoiding the confound of evaluating code quality in a language the model rarely produces. The resulting format distribution is itself an informative signal, analysed in Section~\ref{sec:exp-format}.

\subsubsection{Compilation.}
The output format is auto-detected from language tags and content heuristics. \textbf{TikZ/\LaTeX} (\verb|\begin{tikzpicture}| etc.)\ is compiled with \texttt{pdflatex} (60-second timeout) and converted to a 300-DPI PNG via \texttt{pdftoppm}. A minimal \LaTeX{} preamble is injected if the response contains only a \texttt{tikzpicture} environment without a surrounding document class. \textbf{SVG} (\texttt{<svg} tag) is rendered to $800 \times 800$ pixels via \texttt{cairosvg}. \textbf{Python/Matplotlib} (\texttt{import matplotlib} or \texttt{plt.savefig}) runs in an isolated subprocess with a 30-second timeout, network access disabled, and filesystem writes restricted to the output directory. If the first detected format fails, the pipeline tries the remaining two compilers in sequence. The two text-to-image image models return raster images directly, bypassing compilation entirely.

\subsubsection{Checkpointing and Reproducibility.}
Responses are checkpointed to \path{<model>/responses.json} after every image; compiled PNGs are cached at \path{<model>/<id>.png}. Format distribution, compile rates, and error diagnostics are logged to \path{generation_log.json}, enabling any interrupted run to resume without repeating API calls. Compilation and evaluation run on a CPU-only machine with 32~GB RAM; no GPU is required since all metrics operate on pre-rendered 300-DPI PNGs.

\begin{figure}[!htb]
  \centering
  \includegraphics[width=0.85\textwidth]{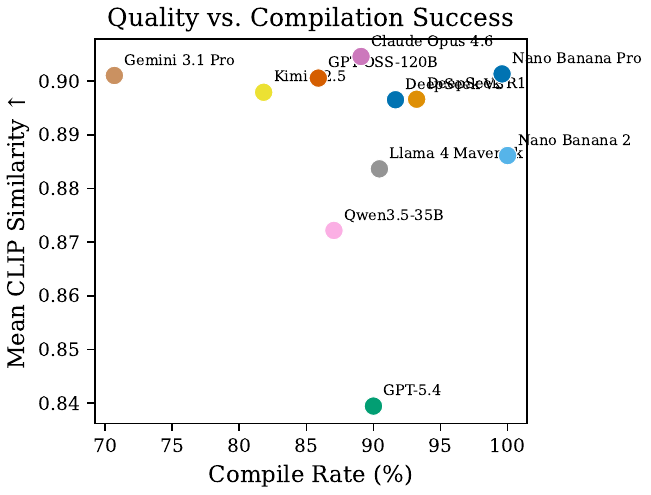}
  \caption{\textbf{Quality vs. compilation success across all models.} Each point represents one model, with compile rate on the $x$-axis and mean CLIP similarity on the $y$-axis. Nano Banana Pro occupies the upper-right region, combining the highest compile reliability with one of the strongest CLIP scores. Nano Banana 2 also achieves near-perfect compile rate, but with noticeably lower CLIP similarity than the top-quality models. Claude Opus 4.6 appears to achieve the highest CLIP similarity overall, while Gemini 3.1 Pro attains strong quality despite a substantially lower compile rate. GPT-5.4 is a clear outlier: it compiles frequently but has the lowest CLIP similarity in the set.}
  \label{fig:scatter_quality}
\end{figure}

\subsection{Error Analysis}
\label{sec:exp-errors}

To move beyond aggregate metrics, we conducted a systematic qualitative analysis of failure modes by manually inspecting 200 generated diagrams sampled across all models, stratified by metric quartile (50 from the best quartile, 100 from the middle, 50 from the worst). We identify five recurring failure categories, ordered by frequency of occurrence.

\subsubsection{Label Hallucination and Omission.}
The most common failure mode across both code and image models is incorrect handling of textual labels. Models frequently hallucinate labels not present in the ground truth (e.g., adding coordinate values to vertices that were unlabelled in the original), omit labels that are present, or place labels at incorrect positions relative to their corresponding geometric elements. This failure is particularly damaging because a single misplaced label can render a mathematical diagram pedagogically misleading.

\subsubsection{Coordinate and Scale Miscalculation.}
Code-generating models frequently produce syntactically valid code with incorrect coordinate arithmetic. Common manifestations include: triangles with vertices at wrong positions that yield incorrect side-length ratios, circles with radii that do not match the specified tangency constraints, and function plots with axes scaled incorrectly relative to the domain specified in the prompt. These errors are insidious because they survive compilation and the code runs and produces a clean-looking diagram that is geometrically wrong. This failure mode is most pronounced in GPT-5.4, which tends to generate elaborate coordinate computations that accumulate rounding or logical errors.

\subsubsection{Topological Errors in Complex Diagrams.}
For diagrams involving graph structures, Venn diagrams, or combinatorial arrangements, models frequently produce outputs with incorrect connectivity. Edges may connect wrong vertices, set intersections may be rendered as disjoint regions, or tree structures may have incorrect branching. These errors are difficult to detect with perceptual metrics (CLIP scores remain high because the ``type'' of diagram is correct) but are immediately apparent to a human reader, underscoring the importance of edge-based structural metrics.

\subsubsection{3D Projection Failures.}
Solid geometry diagrams expose a systematic weakness: models struggle to render consistent 3D projections. Common issues include: inconsistent vanishing points, edges that should be occluded (dashed) rendered as solid, and faces with incorrect relative sizing. This category has the lowest mean CLIP scores and correlates strongly with difficulty level~5 problems.

\subsubsection{Compilation Failures from Over-Complexity.}
Among code models, compilation failures are not random, they concentrate on prompts that require more complex diagrams. Gemini~3.1~Pro's 29.3\% failure rate is dominated by overly ambitious SVG output that exceeds renderer memory limits or contains malformed path definitions. GPT-5.4 produces TikZ code with deeply nested loops and conditional structures that trigger \texttt{pdflatex} timeouts. In contrast, models like DeepSeek~V3 and Claude Opus~4.6 tend to produce simpler, more robust code structures, sacrificing some expressiveness for reliability.

\subsubsection{Common-Subset Analysis.}
\label{sec:exp-common}

Compile rates vary widely, from 70.7\% (Gemini) to 100\% (Nano~Banana~2), so comparing raw averages risks conflating image quality with selection effects: a model that fails on hard prompts appears artificially strong on the ones it manages to complete. To control for this confound, we recomputed all metrics on the \textbf{common subset} of 1,068 images for which every model produced a valid output (Table~\ref{tab:leaderboard_common}).

\begin{table}[!htb]
\centering
\caption{Model performance on the common subset ($n = 1{,}068$ images where all eleven models compiled successfully). Bold = best, underline = second best.}
\label{tab:leaderboard_common}
\small
\begin{tabular}{l r r r r}
\toprule
Model & DISTS\,$\downarrow$ & CLIP\,$\uparrow$ & Edge IoU\,$\uparrow$ & Edge F1\,$\uparrow$ \\
\midrule
Claude Opus~4.6   & \textbf{0.245\,\scriptsize$\pm$.004} & \textbf{0.920\,\scriptsize$\pm$.003} & \underline{0.094\,\scriptsize$\pm$.005} & \underline{0.163\,\scriptsize$\pm$.007} \\
Gemini~3.1~Pro    & \underline{0.249\,\scriptsize$\pm$.006} & \underline{0.918\,\scriptsize$\pm$.004} & \textbf{0.100\,\scriptsize$\pm$.005} & \textbf{0.171\,\scriptsize$\pm$.008} \\
GPT-OSS-120B      & 0.255\,{\scriptsize$\pm$.004} & 0.914\,{\scriptsize$\pm$.003} & 0.072\,{\scriptsize$\pm$.004} & 0.130\,{\scriptsize$\pm$.005} \\
Kimi K2.5         & 0.256\,{\scriptsize$\pm$.005} & 0.915\,{\scriptsize$\pm$.004} & 0.083\,{\scriptsize$\pm$.005} & 0.145\,{\scriptsize$\pm$.007} \\
DeepSeek V3       & 0.251\,{\scriptsize$\pm$.004} & 0.913\,{\scriptsize$\pm$.003} & 0.082\,{\scriptsize$\pm$.004} & 0.146\,{\scriptsize$\pm$.006} \\
DeepSeek R1       & 0.262\,{\scriptsize$\pm$.004} & 0.910\,{\scriptsize$\pm$.003} & 0.076\,{\scriptsize$\pm$.004} & 0.135\,{\scriptsize$\pm$.006} \\
Nano Banana Pro   & 0.274\,{\scriptsize$\pm$.004} & 0.909\,{\scriptsize$\pm$.004} & 0.070\,{\scriptsize$\pm$.003} & 0.129\,{\scriptsize$\pm$.004} \\
Nano Banana 2     & 0.276\,{\scriptsize$\pm$.004} & 0.894\,{\scriptsize$\pm$.004} & 0.071\,{\scriptsize$\pm$.003} & 0.130\,{\scriptsize$\pm$.004} \\
Llama 4 Maverick  & 0.297\,{\scriptsize$\pm$.006} & 0.892\,{\scriptsize$\pm$.004} & 0.058\,{\scriptsize$\pm$.003} & 0.106\,{\scriptsize$\pm$.005} \\
Qwen3.5-35B       & 0.318\,{\scriptsize$\pm$.008} & 0.884\,{\scriptsize$\pm$.005} & 0.060\,{\scriptsize$\pm$.003} & 0.107\,{\scriptsize$\pm$.005} \\
GPT-5.4           & 0.366\,{\scriptsize$\pm$.009} & 0.858\,{\scriptsize$\pm$.006} & 0.043\,{\scriptsize$\pm$.003} & 0.080\,{\scriptsize$\pm$.004} \\
\bottomrule
\end{tabular}
\end{table}

The rankings are largely stable. Claude Opus~4.6 retains the lead on DISTS (0.245) and CLIP (0.920); Gemini~3.1~Pro still wins Edge~IoU (0.100) and Edge~F1 (0.171). The gap between these two and the rest actually \emph{widens} on the common subset, suggesting that their advantage is not an artefact of easy images, it holds on the prompts that every model could handle. GPT-5.4 remains the weakest code model (DISTS~0.366, CLIP~0.858), confirming that its poor showing is not driven by compilation failures on difficult prompts. The Nano~Banana models drop slightly in rank on structural metrics (Edge~IoU~$\approx$~0.070--0.071), indicating that on the subset of prompts all code models can compile, code-based generation holds a modest structural edge.

\begin{figure}[!htb]
  \centering
  \includegraphics[width=\textwidth]{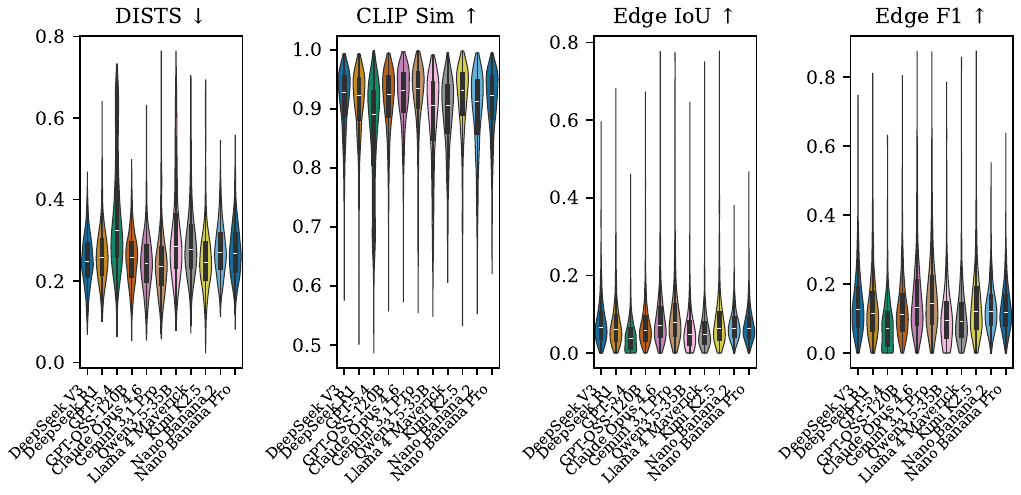}
  \caption{Per-metric score distributions across models. Claude Opus~4.6 and Gemini~3.1~Pro show tighter, more favourable distributions on DISTS and CLIP. GPT-5.4 exhibits a heavy right tail on DISTS, reflecting a subset of severely degraded outputs. Edge~IoU and Edge~F1 distributions are right-skewed for all models, indicating that most diagrams capture only a fraction of the reference edge structure.}
  \label{fig:distributions}
\end{figure}

\begin{figure}[!htb]
  \centering
  \includegraphics[width=0.9\textwidth]{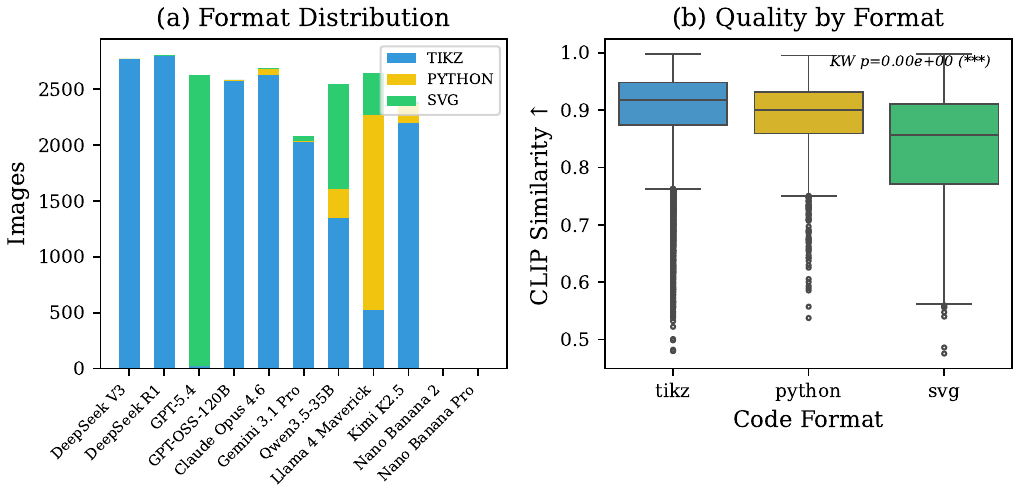}
  \caption{Output format distribution and per-format metric performance across code-generating models. TikZ is the most commonly chosen format and achieves the strongest edge metrics on average.}
  \label{fig:format_analysis}
\end{figure}

Figure~\ref{fig:format_analysis} breaks down performance by output format among the code models. TikZ is the dominant format choice across most models and tends to produce the best structural scores when it compiles, consistent with TikZ's explicit geometric instruction set. Python/Matplotlib is the second most common choice and achieves competitive CLIP scores but lower edge overlap, likely due to the rasterisation artefacts introduced by Matplotlib's anti-aliasing and default styling.

\begin{figure}[!htb]
  \centering
  \includegraphics[width=0.85\textwidth]{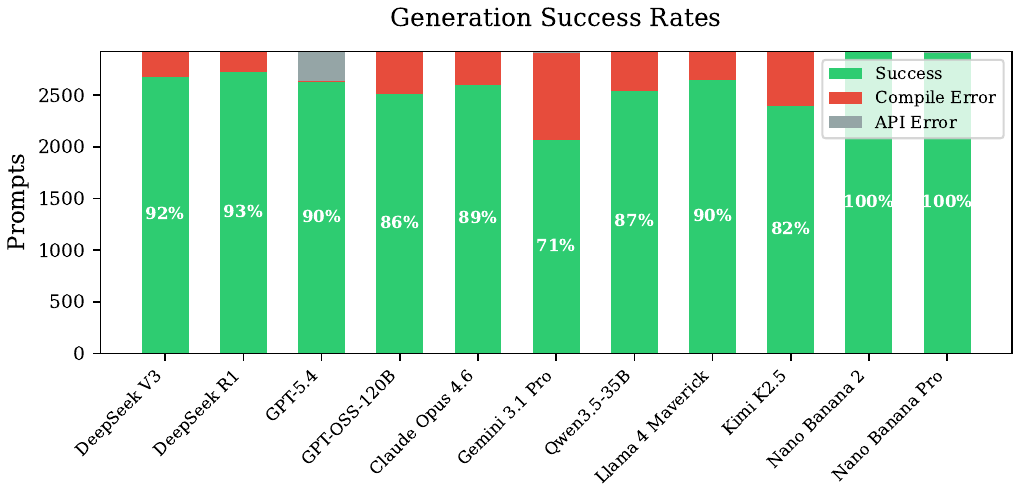}
  \caption{\textbf{Generation success rates across models.} Stacked bars decompose outcomes into successful generations, compile errors, and API errors. Nano Banana 2 and Nano Banana Pro attain 100\% success, reflecting their text-to-image generation setup, which avoids the compilation stage. Among code-generation models, DeepSeek R1 performs best with a 93\% success rate, followed closely by DeepSeek V3 at 92\%, whereas Gemini 3.1 Pro performs worst at 71\%. Unlike the other code models, GPT-5.4 exhibits API failures rather than compile errors.}
  \label{fig:compile_rates}
\end{figure}

\subsubsection{Code Generation vs.\ Direct Image Synthesis.}
\label{sec:exp-format}

A central question in our benchmark design is whether code-based generation (TikZ, SVG, Python) or direct image synthesis better serves mathematical diagram reconstruction. The results paint a nuanced picture.

On \textbf{reliability}, the image models win decisively: Nano~Banana~2 and Pro achieve 100\% and 99.6\% success rates respectively, since there is no compilation step to fail. The best code model, DeepSeek~R1, manages 93.2\%, still a gap of 7 percentage points.

On \textbf{perceptual quality}, the two paradigms are surprisingly close. Nano~Banana~Pro's CLIP score (0.901) ties with Gemini~3.1~Pro and GPT-OSS-120B, and its DISTS (0.283) sits squarely in the middle of the code-model range.

On \textbf{structural precision}, however, the best code models pull ahead. Gemini~3.1~Pro and Claude Opus~4.6 achieve Edge~IoU scores of 0.095 and 0.089 respectively, compared with 0.077--0.079 for the Nano~Banana models. Code-based output explicitly encodes geometric primitives, coordinates, angles, path commands, which the compiler renders faithfully when the code is correct. Image models must infer this structure implicitly and tend to produce softer, less geometrically exact edges.

\subsubsection{Category-Level Analysis.}
\label{sec:exp-categories}

\begin{figure}[!htb]
  \centering
  \includegraphics[width=\textwidth]{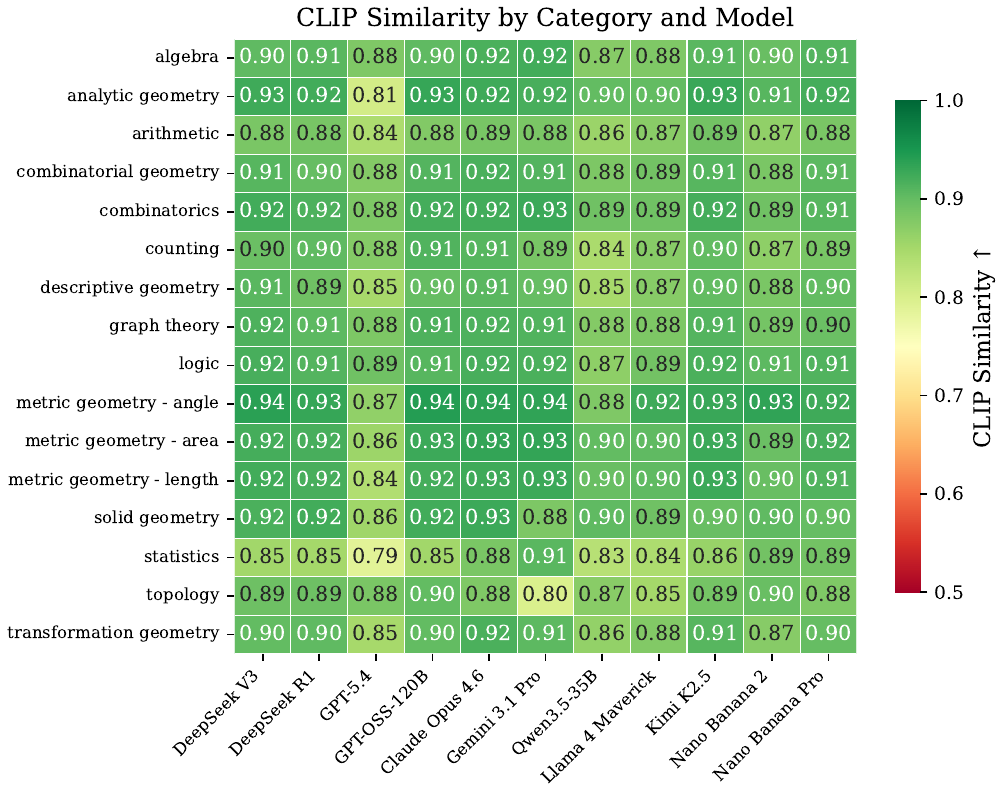}
  \caption{Model $\times$ category heatmap (CLIP similarity, common subset). Darker cells indicate stronger performance. Claude Opus~4.6 and Gemini~3.1~Pro dominate most categories, with Nano~Banana~Pro competitive on geometry but weaker on charts and statistics.}
  \label{fig:category_heatmap}
\end{figure}

Not all mathematical diagrams are equally difficult to reconstruct. Table~\ref{tab:categories} reports per-category averages across all models on the common subset, and Figure~\ref{fig:category_heatmap} gives the full model $\times$ category breakdown.

\begin{table}[!htb]
\centering
\caption{Per-category performance averaged across all models on the common subset, sorted by CLIP similarity. Categories at the top are easiest for current models; those at the bottom are hardest.}
\label{tab:categories}
\small
\begin{tabular}{l r r r r r}
\toprule
Category & $n$ & DISTS\,$\downarrow$ & CLIP\,$\uparrow$ & Edge IoU\,$\uparrow$ & Edge F1\,$\uparrow$ \\
\midrule
Metric geometry, angle     & 39  & 0.257 & 0.922 & 0.069 & 0.124 \\
Metric geometry, area      & 188 & 0.264 & 0.912 & 0.072 & 0.127 \\
Combinatorics               & 72  & 0.278 & 0.909 & 0.079 & 0.137 \\
Analytic geometry            & 48  & 0.290 & 0.908 & 0.068 & 0.123 \\
Metric geometry, length    & 181 & 0.281 & 0.908 & 0.066 & 0.117 \\
Logic                        & 51  & 0.259 & 0.905 & 0.078 & 0.139 \\
Solid geometry               & 34  & 0.280 & 0.902 & 0.083 & 0.144 \\
Algebra                      & 125 & 0.281 & 0.901 & 0.076 & 0.137 \\
Graph theory                 & 41  & 0.286 & 0.900 & 0.083 & 0.149 \\
Combinatorial geometry       & 124 & 0.275 & 0.899 & 0.079 & 0.138 \\
Transformation geometry      & 50  & 0.292 & 0.891 & 0.063 & 0.115 \\
Descriptive geometry         & 23  & 0.274 & 0.888 & 0.074 & 0.134 \\
Counting                     & 13  & 0.308 & 0.887 & 0.065 & 0.118 \\
Topology                     & 5   & 0.314 & 0.876 & 0.068 & 0.125 \\
Arithmetic                   & 44  & 0.301 & 0.875 & 0.082 & 0.147 \\
Statistics                   & 30  & 0.280 & 0.859 & 0.081 & 0.147 \\
\bottomrule
\end{tabular}
\end{table}

The easiest categories are \textbf{metric geometry, angle} (CLIP~0.922) and \textbf{metric geometry, area} (CLIP~0.912). These typically involve simple polygons with labelled vertices and angle arcs, structures well represented in LLM training data and straightforward to express in TikZ.

The hardest categories are \textbf{statistics} (CLIP~0.859) and \textbf{topology} (CLIP~0.876), though the latter has only 5 images in the common subset. Statistical charts require precise axis scaling, data-point placement, and legend rendering that code models handle unevenly. Topological diagrams, knots, surface cross-sections, homeomorphic mappings, demand spatial reasoning about non-standard shapes that current models rarely encounter during training.

\textbf{Transformation geometry} (CLIP~0.891) and \textbf{arithmetic} (CLIP~0.875) also prove challenging. Transformation diagrams involve reflection, rotation, and scaling operations that must be composed correctly, while arithmetic diagrams (e.g.\ number lines, base-conversion visuals) require exact numeric placement that models frequently get wrong.

\begin{figure}[!htb]
  \centering
  \includegraphics[width=0.85\textwidth]{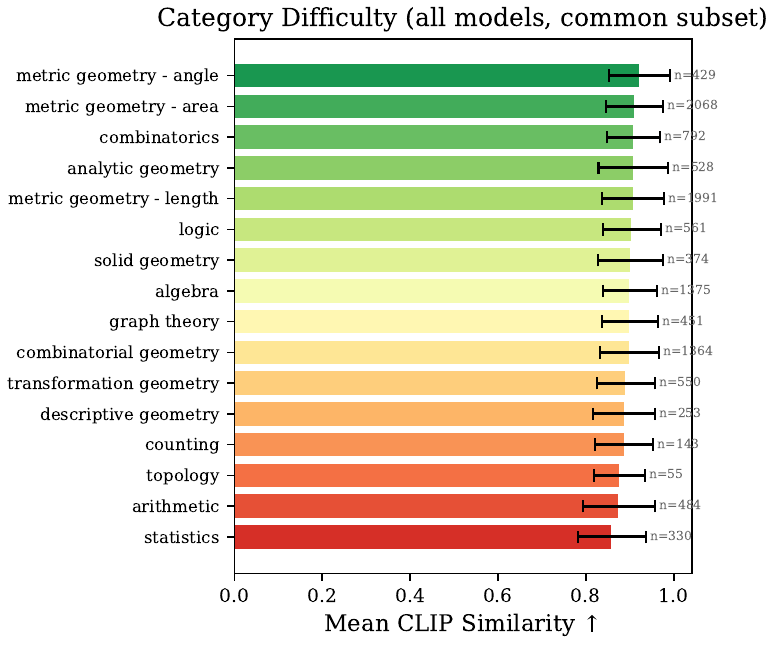}
  \caption{Mean CLIP similarity as a function of MathVision difficulty level (1 = easiest, 5 = hardest). All models degrade with increasing difficulty, but the slope is relatively uniform across model families.}
  \label{fig:difficulty}
\end{figure}

\FloatBarrier

\subsubsection{Statistical Significance.}
\label{sec:exp-significance}

\begin{figure}[!htb]
  \centering
  \includegraphics[width=0.85\textwidth]{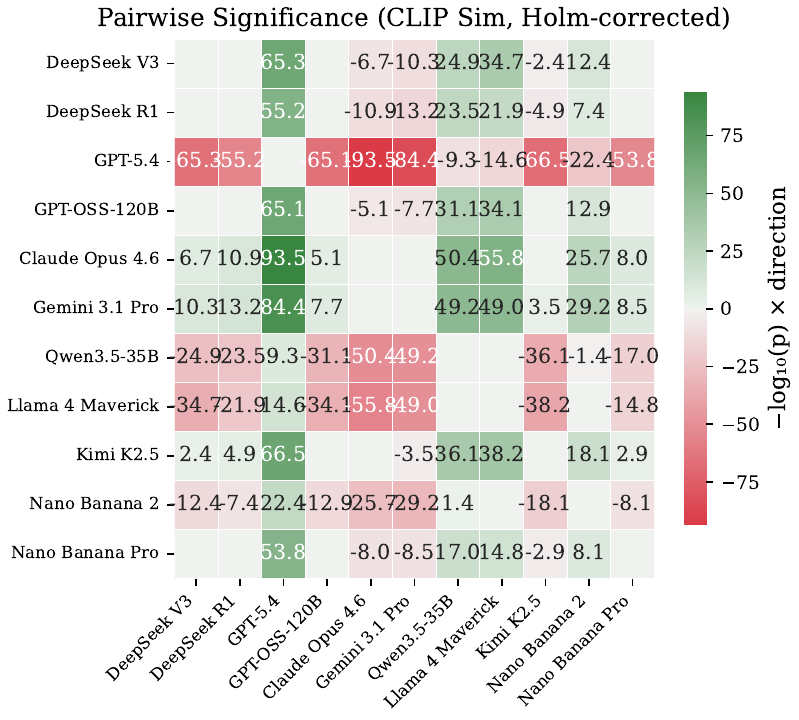}
  \caption{Pairwise Wilcoxon signed-rank significance matrix (Holm--Bonferroni corrected, $\alpha = 0.05$). Filled cells indicate statistically significant differences between the corresponding model pair.}
  \label{fig:significance}
\end{figure}

With eleven models and four per-image metrics, it is important to verify that observed differences are not due to chance. Figure~\ref{fig:significance} presents the pairwise significance matrix.

The top-tier cluster, Claude Opus~4.6, Gemini~3.1~Pro, GPT-OSS-120B, and DeepSeek~V3, separates significantly from the mid-tier models (Kimi~K2.5, Llama~4~Maverick, Qwen3.5-35B) on most metrics. Within the top tier, Claude and Gemini differ significantly on edge metrics (Gemini ahead) but not on CLIP. GPT-5.4 is significantly worse than every other model on all four metrics. The Nano~Banana models form a distinct group: their DISTS and CLIP scores are not significantly different from mid-tier code models, but their edge scores are significantly lower than Claude and Gemini, confirming the structural-precision gap.

\subsubsection{Metric Correlations.}
\label{sec:exp-correlations}

\begin{figure}[!htb]
  \centering
  \includegraphics[width=\textwidth]{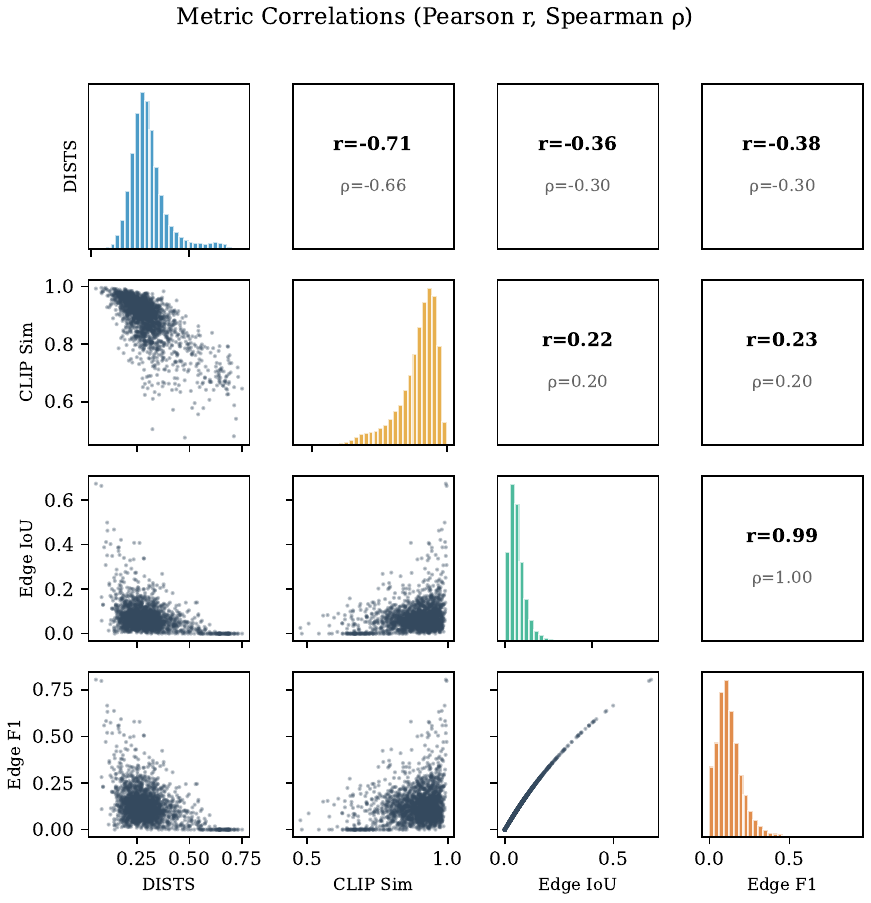}
  \caption{Pairwise scatter plots and Pearson correlations between the four per-image metrics, pooled across all models. DISTS and CLIP are strongly (anti-)correlated; edge metrics are nearly redundant with each other but only moderately correlated with the perceptual metrics.}
  \label{fig:correlations}
\end{figure}

Figure~\ref{fig:correlations} plots the pairwise correlations between our four per-image metrics. DISTS and CLIP similarity are strongly negatively correlated ($r \approx -0.7$), which is expected: both capture perceptual/semantic similarity from opposite ends of the scale. Edge~IoU and Edge~F1 are nearly perfectly correlated ($r > 0.95$), reflecting the mathematical relationship between IoU and Dice.

More informative is the moderate correlation between CLIP and the edge metrics ($r \approx 0.4$--$0.5$). This confirms that semantic similarity and structural overlap are related but not redundant: a model can produce an image that CLIP considers semantically correct yet still miss fine geometric detail, and vice versa. The partial independence of these metric families validates our decision to evaluate along both axes rather than relying on a single score.

\FloatBarrier

\subsection{Summary of Findings}
\label{sec:exp-summary}

We distill the experimental results into five key takeaways:

\begin{enumerate}
  \item \textbf{No single code-generating model dominates all axes.} Claude Opus~4.6 leads on perceptual/semantic metrics; Gemini~3.1~Pro leads on structural precision; Nano~Banana~2 leads on reliability. Model selection depends on which property matters most for a given application.

  \item \textbf{Compilation reliability is a first-order concern.} Gemini~3.1~Pro produces the best diagrams \emph{when it compiles}, but loses nearly 30\% of its outputs. In a production pipeline, the combination of high quality and low reliability may be worse than a model with moderate quality and high reliability.

  \item \textbf{Code-based generation retains a structural advantage} over direct image synthesis, particularly on geometric detail. The Nano~Banana models are perceptually competitive but fall short on edge overlap, suggesting that explicit geometric primitives in code still matter for mathematical precision.

  \item \textbf{Mathematical subject strongly modulates difficulty.} Models handle standard plane-geometry diagrams well but struggle with statistical charts, topological figures, and transformation diagrams. Future work on diagram generation should target these underperforming categories specifically.

  \item \textbf{GPT-5.4's poor showing warrants investigation.} The flagship OpenAI model is consistently the weakest code generator in our benchmark, a result that is statistically significant and persists on the common subset. We hypothesize that its extended chain-of-thought reasoning, while beneficial for problem solving, may actively hurt code generation by producing overly complex output.
\end{enumerate}

\subsection{Evaluation Protocol}
\label{sec:exp-metrics}

No single metric captures all dimensions of diagram similarity: pixel agreement misses structural correctness, and semantic similarity can overlook local geometric errors. We therefore employ four per-image metrics that span complementary levels of visual abstraction.

\paragraph{DISTS : Deep Image Structure and Texture Similarity ($\downarrow$).}
DISTS is a full-reference perceptual metric computed on a VGG feature hierarchy. Its invariance to texture exchange is particularly appropriate for mathematical diagrams: a diagram with thicker strokes or different shading should not be penalized if the geometry is correct. Unlike SSIM or LPIPS, DISTS exhibits this invariance by design.

\paragraph{CLIP Cosine Similarity ($\uparrow$).}
We embed both the generated and reference images with CLIP ViT-B/32 and compute cosine similarity. CLIP features represent what an image conceptually \emph{shows}, capturing whether the model produced the correct diagram type and arrangement even when the visual rendering differs from the reference.

\paragraph{Edge IoU and Edge F1 ($\uparrow$).}
Mathematical diagrams are fundamentally line drawings whose geometric content resides in edges, not fills. We extract binary edge maps with the Canny detector ($3 \times 3$ dilation) and compute:
\[
  \text{Edge IoU} = \frac{|E_{\text{gen}} \cap E_{\text{ref}}|}
                        {|E_{\text{gen}} \cup E_{\text{ref}}|},
  \qquad
  \text{Edge F1}  = \frac{2|E_{\text{gen}} \cap E_{\text{ref}}|}
                        {|E_{\text{gen}}| + |E_{\text{ref}}|}.
\]
Both measure geometric skeleton overlap, insensitive to background, fill colour, or minor styling differences.

\paragraph{Limitations}
This benchmark focuses on static 2D diagrams with English-only prompts. The edge metrics (IoU and F1) operate at the floor of their range (best model achieves Edge~IoU of 0.100), raising questions about whether Canny-based edge detection with fixed parameters is sufficiently discriminative across diagrams with varying stroke widths and rendering styles; future work should explore adaptive edge extraction or learned structural metrics. This begs the need for a new set of appropriate metrics to compare the mathematical validity of two diagrams. We also experimented with CMMD (CLIP Maximum Mean Discrepancy), a distributional metric over CLIP embeddings, but it could not be meaningfully compared across models due to the unequal number of valid pairs per model and is therefore omitted from the per-model tables; we note it here for future iterations where common-subset CMMD computation is feasible. We recommend future work on metrics, comprehensiveness of diagram classes and pipeline improvement.

\section{Full SME Prompt-Sufficiency Scores}
\label{sec:sme-appendix}

Table~\ref{tab:sme-full} lists the complete per-image scores from both SMEs over the 50-pair validation sample described in Section~\ref{sec:prompt-sufficiency}. Scores are on a 1--5 scale for Completeness (Comp.), Correctness (Corr.), and Clarity (Clar.).

\begin{table}[!htb]
\centering
\caption{Full SME prompt-sufficiency scores ($n=50$). Comp./Corr./Clar.\ follow the rubric in Section~\ref{sec:prompt-sufficiency}.}
\label{tab:sme-full}
\scriptsize
\setlength{\tabcolsep}{4pt}
\begin{tabular}{r l ccc ccc}
\toprule
& & \multicolumn{3}{c}{\textbf{SME 1}} & \multicolumn{3}{c}{\textbf{SME 2}} \\
\cmidrule(lr){3-5}\cmidrule(lr){6-8}
\textbf{ID} & \textbf{Category} & Comp. & Corr. & Clar. & Comp. & Corr. & Clar. \\
\midrule
2998 & metric geometry - area & 5 & 5 & 5 & 5 & 5 & 5 \\
169 & combinatorial geometry & 5 & 5 & 5 & 5 & 5 & 5 \\
322 & metric geometry - length & 5 & 5 & 5 & 5 & 5 & 5 \\
2186 & metric geometry - area & 5 & 5 & 4 & 5 & 5 & 5 \\
2506 & algebra & 5 & 5 & 5 & 5 & 5 & 5 \\
2954 & metric geometry - angle & 5 & 5 & 5 & 5 & 4 & 5 \\
3036 & metric geometry - area & 5 & 5 & 5 & 5 & 5 & 5 \\
1005 & combinatorial geometry & 5 & 5 & 4 & 5 & 5 & 4 \\
509 & arithmetic & 5 & 3 & 3 & 4 & 3 & 3 \\
2532 & solid geometry & 5 & 5 & 5 & 5 & 5 & 5 \\
2737 & metric geometry - area & 5 & 5 & 5 & 5 & 5 & 5 \\
1821 & algebra & 5 & 5 & 5 & 5 & 5 & 5 \\
1157 & algebra & 5 & 5 & 5 & 5 & 5 & 5 \\
2936 & metric geometry - area & 5 & 5 & 5 & 4 & 5 & 4 \\
2407 & metric geometry - area & 5 & 5 & 5 & 5 & 5 & 5 \\
2952 & metric geometry - length & 5 & 5 & 5 & 5 & 5 & 5 \\
2605 & solid geometry & 5 & 5 & 4 & 4 & 5 & 3 \\
2034 & logic & 5 & 5 & 5 & 5 & 5 & 5 \\
2459 & metric geometry - length & 5 & 5 & 4 & 5 & 4 & 5 \\
2106 & combinatorics & 5 & 5 & 5 & 5 & 5 & 5 \\
3003 & metric geometry - area & 5 & 5 & 5 & 5 & 5 & 5 \\
3031 & metric geometry - angle & 5 & 5 & 5 & 5 & 5 & 5 \\
2681 & metric geometry - area & 5 & 5 & 5 & 5 & 5 & 5 \\
2047 & algebra & 5 & 5 & 5 & 5 & 5 & 5 \\
2749 & metric geometry - area & 5 & 5 & 5 & 5 & 5 & 5 \\
2667 & combinatorial geometry & 5 & 5 & 5 & 5 & 5 & 5 \\
1238 & logic & 5 & 5 & 5 & 5 & 5 & 5 \\
1273 & arithmetic & 5 & 5 & 5 & 5 & 5 & 5 \\
146 & logic & 5 & 5 & 5 & 5 & 4 & 4 \\
706 & combinatorics & 5 & 4 & 4 & 5 & 5 & 5 \\
2026 & algebra & 5 & 5 & 5 & 5 & 5 & 5 \\
2022 & combinatorics & 5 & 5 & 5 & 5 & 5 & 5 \\
905 & algebra & 5 & 4 & 3 & 4 & 4 & 4 \\
2625 & combinatorial geometry & 5 & 5 & 5 & 5 & 5 & 4 \\
1159 & transformation geometry & 5 & 5 & 5 & 5 & 4 & 5 \\
691 & combinatorics & 5 & 5 & 5 & 5 & 5 & 5 \\
2028 & algebra & 5 & 5 & 5 & 5 & 5 & 5 \\
2228 & solid geometry & 5 & 4 & 4 & 5 & 4 & 4 \\
2131 & combinatorics & 5 & 4 & 5 & 5 & 4 & 5 \\
2090 & solid geometry & 5 & 5 & 5 & 5 & 5 & 5 \\
841 & graph theory & 5 & 5 & 5 & 5 & 5 & 5 \\
1656 & logic & 5 & 5 & 5 & 5 & 5 & 5 \\
2244 & combinatorics & 5 & 5 & 5 & 4 & 5 & 4 \\
483 & combinatorics & 5 & 5 & 5 & 5 & 5 & 5 \\
2300 & metric geometry - length & 5 & 5 & 5 & 5 & 5 & 5 \\
2232 & combinatorics & 5 & 5 & 5 & 5 & 5 & 5 \\
1599 & combinatorial geometry & 5 & 5 & 5 & 5 & 5 & 5 \\
1290 & combinatorics & 5 & 5 & 5 & 5 & 5 & 5 \\
2119 & metric geometry - area & 5 & 5 & 5 & 5 & 5 & 4 \\
1329 & logic & 5 & 5 & 5 & 5 & 5 & 5 \\
\midrule
\multicolumn{2}{l}{\textbf{Mean}} & 5.00 & 4.88 & 4.80 & 4.90 & 4.82 & 4.76 \\
\bottomrule
\end{tabular}
\end{table}

\section{Discussion}

Math-Vision Diagrams reveals persistent challenges in mathematical diagram generation. Text-to-code models excel in structural and semantic tasks due to their symbolic nature, while text-to-image models suffer from hallucinated labels and imprecise geometries despite strong perceptual quality. Limitations include focus on static 2D diagrams and English-only prompts. Researchers can also plan to expand the benchmark with dynamic visualizations and real-time generation scenarios. Given that the benchmark must be sacrosanct, we are open-sourcing all the code, results so that the community can identify potentially any errors or contribute towards its improvements.
These findings have direct practical implications. For educational content pipelines where mathematical correctness is paramount, code-based generation with compilation verification provides the strongest guarantees. For rapid prototyping or contexts where approximate visual fidelity suffices, text-to-image models offer superior reliability. Hybrid approaches, code generation with image-model fallback on compilation failure, represent a promising direction for production systems. Future work will extend the benchmark to interactive and animated diagrams, multilingual prompts, integration with formal theorem provers for constraint-verified evaluation, and improved LLM-as-a-judge calibration protocols to narrow the gap between automated and human expert assessments.

\end{document}